\documentclass[letterpaper]{article} 

\usepackage[preprint]{aaai2027} 
\usepackage[hyphens]{url} 
\usepackage{graphicx} 
\usepackage{natbib} 
\usepackage{caption} 
\usepackage{amsthm} 
\newtheorem{definition}{Definition}
\usepackage{algorithm}
\usepackage{algorithmic}
\usepackage{amsthm}
\usepackage{booktabs}
\usepackage{newfloat}
\usepackage{listings}
\DeclareCaptionStyle{ruled}{labelfont=normalfont,labelsep=colon,strut=off} 
\floatstyle{ruled}
\newfloat{listing}{tb}{lst}{}
\floatname{listing}{Listing}

\usepackage{booktabs}

\usepackage{amsmath}
\usepackage{bm}
\usepackage{amsthm}
\usepackage{amssymb}

\definecolor{customcyan}{RGB}{10, 204, 0} 
\definecolor{tealblue}{RGB}{0, 132, 194}
\definecolor{darkorange}{RGB}{220, 100, 0} 

\title{Benchmarking Sheaf Neural Networks for Inductive Tasks}

\author{
  Stefano Fiorini\textsuperscript{\rm 1}\equalcontrib\corresponding,
  Edoardo Coppola\textsuperscript{\rm 2}\equalcontrib\corresponding,
  Pietro Liò\textsuperscript{\rm 2}
}

\affiliations{
  \textsuperscript{\rm 1}Independent Researcher\\
  \textsuperscript{\rm 2}Department of Computer Science and Technology, University of Cambridge\\
  S.fiorini1994@gmail.com, ec2013@cam.ac.uk, pl219@cam.ac.uk
}

\begin{document}

\maketitle

\begin{abstract}

Sheaf Neural Networks (SNNs) generalize message passing by replacing scalar edge weights of standard Graph Neural Networks (GNNs) with learnable, edge-dependent restriction maps between node stalks. Despite their strong theoretical foundations and promising transductive results, SNNs have been evaluated almost exclusively on transductive node classification, leaving their behaviour under inductive protocols unknown. 
We address this gap through the first systematic benchmark of the sheaf design space, evaluating three diffusion mechanisms (neural sheaf diffusion, sheaf attention, and sheaf attention with Graph Attention Network v2), three restriction-map parameterizations, three stalk dimensions, and six modern GNN architectural components, within a message-passing reformulation that never assembles the heavy sheaf Laplacian, making the full design space trainable under cross-graph batching. 
Across $1{,}890$ controlled experiments on 14 inductive datasets, multiple insights emerge: restriction maps are the dominant design choice and general maps are preferable, larger stalks add capacity but not long-range reach, architectural components explain more performance variation than the entire sheaf-specific design space itself. Under a matched protocol, SNNs transfer to inductive settings but do not reach the strongest baselines, with gaps being dataset-dependent. Practically, a single sheaf configuration can generalize across datasets, so effort is better spent tuning the surrounding architectural recipe than the sheaf operator itself.
\end{abstract}


\section{Introduction}

\begin{figure}[htb!]
\centering
\includegraphics[width=\columnwidth]{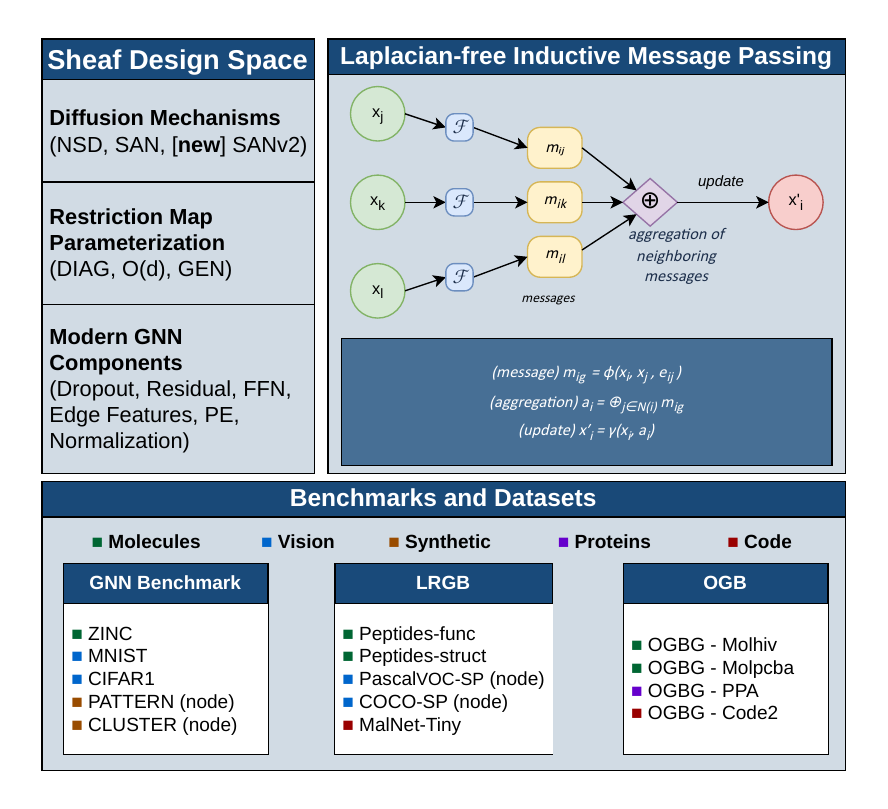} 
\caption{Benchmarking workflow overview. (\textbf{Top-left}) sheaf design space spanning 3 diffusion mechanisms, 3 restriction-map parameterizations, and 6 GNN components; (\textbf{top-right}) our Laplacian-free inductive message passing; (\textbf{bottom}) evaluation benchmarks, datasets, and domain taxonomy.}
\label{fig:main-figure}
\end{figure}

Graph Neural Networks (GNNs) have become the dominant paradigm for learning from graph-structured data, achieving remarkable success across a wide range of applications, including molecular prediction~\cite{abramson2024accurate}, biological networks~\citep{traversa2023robustnesscomplexitydirectedweighted}, and recommender systems~\citep{wu2022graph}. Despite their success, conventional message-passing GNNs suffer from fundamental limitations, including oversmoothing, oversquashing, and reduced expressiveness on heterophilous graphs, motivating the development of increasingly expressive graph learning architectures~\cite{li2018deeper, alon2020bottleneck}. Among these, Sheaf Neural Networks (SNNs) represent one of the most principled extensions of message passing, replacing the classical graph Laplacian with a learnable sheaf Laplacian and propagating information through edge-dependent restriction maps rather than scalar weights~\cite{hansen2020sheaf}. This richer diffusion mechanism has been shown, both theoretically and empirically, to improve robustness to heterophily and mitigate oversmoothing relative to standard message passing~\cite{bodnar2022neural}.

However, despite a rapidly growing body of work on sheaf-based graph learning~\cite{barbero2022sheaf, barbero2022connection, duta2023sheaf, fiorini2025sheaves}, our empirical understanding of these models remains almost exclusively confined to \emph{transductive node classification}, where training and inference are performed on a single, fixed graph. While such benchmarks have historically been central to graph representation learning, they capture only a narrow slice of real-world applications: molecular learning, protein modelling, graph classification, and dynamic networks all require \emph{inductive generalization}, where models must operate on previously unseen graphs or unseen regions of a graph.
Reflecting this, graph learning research has progressively shifted its focus from transductive citation networks toward large-scale inductive benchmarks that better reflect realistic deployment scenarios, a shift that sheaf-based models have yet to follow.

This shift has also highlighted how sensitive architectural comparisons are to the evaluation protocol. Once model selection, encoders, and tuning budgets are standardized, much of the performance gap between classical GNNs and more sophisticated architectures disappears~\citep{errica2019fair, dwivedi2023benchmarking, tonshoff2023did}. In particular, augmenting classical GNNs with modern architectural components enables them to match or even outperform substantially more expensive Graph Transformers (GTs) across 14 inductive benchmarks~\citep{luo2025can}, establishing strong baselines for inductive graph learning.
SNNs have remained largely absent from this re-evaluation, {with prior applications limited to narrow settings~\citep{ribeiro2025cooperative}.}  
More broadly, it remains unclear whether restriction-map type, diffusion mechanism, or their interaction drives inductive performance, and whether sheaf diffusion benefits from the modern architectural components that have proven decisive for classic GNNs. Addressing these gaps requires a systematic evaluation of SNNs under contemporary inductive protocols.

This paper addresses this gap through the first systematic study of SNNs under a modern inductive protocol. Since existing sheaf-diffusion implementations target a single fixed graph and cannot operate on cross-graphs batches, we first reformulate sheaf diffusion as a Laplacian-free, edge-wise message-passing operator trainable across batched graphs. We then systematically evaluate the sheaf design space across three diffusion mechanisms, three restriction-map parameterizations, three stalk dimensions, and leave-one-out ablations of six modern GNN architectural components on 14 graph- and node-level datasets, confirming the selected configurations at full budget. The resulting picture yields both mechanistic insight into what drives inductive sheaf performance and actionable guidance for where future effort should go.

\textbf{Our contributions are summarized as follows:}

\begin{itemize}
  \item An inductive, Laplacian-free, message-passing reformulation of sheaf diffusion that enables the first systematic characterization and evaluation of SNNs under a modern inductive protocol. The same reformulation introduces SANv2, a new sheaf-attention layer with Graph Attention (GAT) v2-style dynamic attention.
  \item A characterization of the sheaf design space over 1,890 controlled experiments, revealing, among others, that restriction-map parameterization is the dominant sheaf-specific design choice, increased stalk dimension adds capacity rather than long-range reach, and the diffusion mechanism has negligible aggregate effect.
  \item Actionable design guidance for inductive SNNs: the surrounding GNN+ architecture explains more performance variation than the entire sheaf design space and a single sheaf configuration generalizes across datasets, so tuning is better spent on the recipe than on the sheaf. We further quantify, dataset by dataset, the remaining gap to the strongest baselines.

\end{itemize}

\section{Preliminaries}
\label{sec:preliminaries}

Given an undirected graph $G=(V,E)$ with $n=|V|$ nodes and $m=|E|$ edges, let $\mathcal{N}(i)=\{j:\{i,j\}\in E\}$ denote the neighborhood of node $i$. We use ${I}_k$ for the $k\times k$ identity, $\otimes$ for the Kronecker product, and $\odot$ for the Hadamard product.

\subsection{Cellular Sheaves and the Sheaf Laplacian}
\label{subsec:sheaf}

\begin{definition}
\label{def:sheaf}
A cellular sheaf $(G,\mathcal{F})$ over an undirected graph $G$ associates a vector space $\mathcal{F}(i)$ (the \emph{node stalk}) to each node $i\in V$, a vector space $\mathcal{F}(e)$ (the \emph{edge stalk}) to each edge $e\in E$, and a linear \emph{restriction map} $\mathcal{F}_{i\trianglelefteq e}:\mathcal{F}(i)\to\mathcal{F}(e)$ for each incident node--edge pair $i\trianglelefteq e$.
\end{definition}

Following~\citet{bodnar2022neural}, we take all stalks isomorphic to $\mathbb{R}^{d}$ for a fixed stalk dimension $d$, so each restriction map is a $d \times d$ matrix. In the opinion-dynamics interpretation~\cite{hansen2020sheaf}, node stalks $\mathcal{F}(i)$ represent a node's private space, and $\mathcal{F}_{i\trianglelefteq e}\bm{x}_{i}$ describes how that private state is expressed in the shared edge space $\mathcal{F}(e)$. The space of $0$-cochains $C^0(G,\mathcal{F})=\bigoplus_{i\in V}\mathcal{F}(i)$ stacks the node-stalk vectors; the space of $1$-cochains $C^1(G,\mathcal{F})=\bigoplus_{e\in E}\mathcal{F}(e)$ stacks the edge-stalk vectors. Fixing an arbitrary orientation $e=i\to j$ per edge, the coboundary operator $\delta:C^0(G,\mathcal{F})\to C^1(G,\mathcal{F})$ is $(\delta{X})_e=\mathcal{F}_{j\trianglelefteq e}\bm{x}_j-\mathcal{F}_{i\trianglelefteq e}\bm{x}_i$, and the sheaf Laplacian is $\bm{L}_{\mathcal{F}}=\delta^{\top}\delta$.

\begin{definition}
\label{def:sheaf-laplacian}
The sheaf Laplacian acts on a $0$-cochain $\bm{X}$ node-wise as
\begin{equation}
\label{eq:sheaf-laplacian}
\bm{L}_{\mathcal{F}}({X})_i =\sum_{i,j \trianglelefteq e} \mathcal{F}_{i\trianglelefteq e}^{\top} \left(\mathcal{F}_{i\trianglelefteq e}\bm{x}_i-\mathcal{F}_{j\trianglelefteq e}\bm{x}_j\right).
\end{equation}
\end{definition}

As a matrix, ${L}_{\mathcal{F}}$ is symmetric positive semi-definite with diagonal blocks ${L}_{ii}=\sum_{i\trianglelefteq e}\mathcal{F}_{i\trianglelefteq e}^{\top}\mathcal{F}_{i\trianglelefteq e}$ and off-diagonal blocks ${L}_{ij}=-\mathcal{F}_{i\trianglelefteq e}^{\top}\mathcal{F}_{j\trianglelefteq e}$. The normalized sheaf Laplacian is ${\Delta}_{\mathcal{F}}={D}^{-1/2}{L}_{\mathcal{F}}{D}^{-1/2}$, with ${D}$ the block diagonal of ${L}_{\mathcal{F}}$. For the trivial sheaf, $d=1$, ${\Delta}_{\mathcal{F}}$ reduces to the normalized graph Laplacian, so the sheaf Laplacian strictly generalizes the operator underlying graph convolution \citep{kipf2017semisupervised}. Its kernel $\ker({L}_{\mathcal{F}})=\{{X}:\mathcal{F}_{i\trianglelefteq e}\bm{x}_i=\mathcal{F}_{j\trianglelefteq e}\bm{x}_j\ \forall e=\{i,j\}\}$ is the space of signals in agreement across every edge, and the asymptotics of sheaf diffusion are governed by it.

\subsection{Neural Sheaf Diffusion}
\label{subsec:nsd}
 
\citet{bodnar2022neural} build Neural Sheaf Diffusion (NSD) by
discretizing the sheaf diffusion PDE $\dot{{X}}(t)=-{\Delta}_{\mathcal{F}}{X}(t)$. After lifting the input features into $h$ channels per stalk dimension, an NSD layer applies
\begin{equation}
\label{eq:nsd}
{X}_{t+1}
={X}_t-\sigma\left(
{\Delta}_{\mathcal{F}(t)}\left({I}_n\otimes{W}_{1,t}\right){X}_t{W}_{2,t}
\right),
\end{equation} 
where $\sigma$ is a point-wise nonlinearity and
${W}_{1,t}\in\mathbb{R}^{d\times d}$, ${W}_{2,t}\in\mathbb{R}^{f\times f}$ mix the stalk and channel dimensions. Crucially, the sheaf $\mathcal{F}(t)$ is itself \emph{learned} and evolves with depth: each restriction map is produced by a parametric function of the incident node features, $\mathcal{F}_{i\trianglelefteq e:=(i,j)}=\Phi(\bm{x}_i,\bm{x}_j)$, with $\Phi$ typically an MLP on the concatenation, i.e. $\left(\bm{x}_i\Vert\bm{x}_j\right)$. \citet{bodnar2022neural} show that, in the infinite-depth limit, sheaf diffusion with sufficiently rich restriction maps can linearly separate classes in heterophilic settings where standard graph diffusion provably oversmooths. The relevant axis is the structure imposed on the
restriction maps.

\subsection{Sheaf Attention}
\label{subsec:sheaf-attention}
 
Sheaf Attention Networks (SAN)~\citep{barbero2022sheaf} augment sheaf diffusion with input-dependent attention, generalizing the GAT architecture~\citep{velivckovic2018graph} to SNNs.
Let ${A}_{\mathcal{F}}$ denote the sheaf adjacency with self-loops, whose $(i,j)$-th block is given by the transport map ${P}_{ij}=\mathcal{F}_{i\trianglelefteq e}^{\top}\mathcal{F}_{j\trianglelefteq e}$. A scalar attention matrix ${\Lambda}$ is computed as in GAT, $\Lambda_{ij}=\mathrm{softmax}_j\!\big(\mathrm{LeakyReLU}({a}^{\top}[{W}{x}_i\Vert{W}{x}_j])\big)$, and lifted to stalk dimension as $\hat{{\Lambda}}={\Lambda}\otimes\bm{1}_d$. The SAN layer is defined as
\begin{equation}
\label{eq:sheaf-attention}
{X}_{t+1}
=\sigma\!\left(
\left(\hat{{\Lambda}}({X}_t)\odot{A}_{\mathcal{F}}\right)
\left({I}_n\otimes{W}_{1,t}\right){X}_t{W}_{2,t}
\right),
\end{equation} 
with a residual variant (Res-SAN) inserting an $(\hat{{\Lambda}}\odot{A}_{\mathcal{F}}-{I})$ update around a skip connection. GAT is recovered when $d=1$ and the restriction maps are equal to $1$. Since ${P}_{ij}$ and $\Lambda_{ij}$ are learned independently, feature transport is decoupled from neighbor weighting. \citet{barbero2022sheaf} parameterize $O(d)$ maps and let attention coefficients determine neighbors relative contributions.

\subsection{Restriction Map Families}
\label{subsec:restriction-maps}

The restriction maps $\mathcal{F}_{i\trianglelefteq e}\in\mathbb{R}^{d\times d}$ can be organized into a hierarchy of increasingly general parameterizations~\citep{bodnar2022neural}:

\begin{description}
 \item[Diagonal (\textsc{Diag}).] $\mathcal{F}_{i\trianglelefteq e}$
 is diagonal, scaling each stalk coordinate independently. This parametrization has fewest
 parameters and stalk dimensions interact only through the left
 multiplication by ${W}_1$.
 \item[Orthogonal (\textsc{O($d$)}).] $\mathcal{F}_{i\trianglelefteq e}\in O(d)$, realizing a \emph{discrete $O(d)$-bundle}. Orthogonal maps mix stalk dimensions while preserving norms, regularize via fewer free parameters ($d(d-1)/2$), and yield Laplacians that are numerically convenient to normalize.
 \item[General (\textsc{Gen}).] $\mathcal{F}_{i\trianglelefteq e}$ is an unconstrained $d\times d$ matrix ensuring maximal flexibility and more parameters at the cost of greater overfitting risk and a Laplacian that is harder to normalize numerically (requiring SVD).
\end{description}

\section{Related Works}
\label{sec:related-works}

\paragraph{Sheaf Neural Networks.}
SNNs equip a graph with a cellular sheaf, attaching vector-space stalks to nodes and edges and linear restriction maps to incident pairs, thereby generalizing the graph Laplacian to a sheaf Laplacian \citep{hansen2019spectral}. While \citet{hansen2020sheaf} introduced the first sheaf network with a hand-crafted Laplacian, \citet{bodnar2022neural} made the approach practical with Neural Sheaf Diffusion (NSD), learning the restriction maps end-to-end and establishing that sheaf diffusion can linearly separate classes in heterophilic regimes where graph diffusion oversmooths. Subsequent work extended NSD along several axes: attention \citep{barbero2022sheaf}, connection Laplacians obtained via Riemannian preprocessing \citep{barbero2022connection}, hypergraphs \citep{duta2023sheaf}, heterogeneous graphs \citep{braithwaite2024heterogeneous}, and directionality \citep{fiorini2025sheaves, ribeiro2025cooperative}. Across this line, evaluation is almost entirely \emph{transductive node classification} on a fixed set of heterophilic benchmarks and, notably, the reported best model family is inconsistent across papers ($O(d)$ maps for \citet{bodnar2022neural}, \textsc{GEN} maps for \citet{braithwaite2024heterogeneous}), indicating the optimal diffusion--map parameterization design is task-dependent and, to date, uncharacterized in inductive settings. 

\paragraph{GNN Benchmarking.}
A parallel literature has shown GNN gains are fragile to evaluation protocol: once model selection and budgets are standardized, much of the reported gap between classic GNNs and sophisticated architectures (e.g. GTs) disappears \citep{errica2019fair, dwivedi2023benchmarking, tonshoff2023did}. Most recently, \citet{luo2025can} pushed this further, showing component-augmented classic GNNs (GCN, GIN, GatedGCN) match or surpass state-of-the-art GTs across 14 inductive graph- and node-level datasets at a fraction of their cost, and establishing the standard evaluation protocol and baselines we adopt.

\section{Sheaf Models and Inductive Implementation}
\label{sec:methods}

Throughout, $d$ is the stalk dimension, $f$ the channels per stalk, $c=df$ the per-node representation size.
Node-level tasks feed the node-stalk representations to a per-node head, while graph-level tasks first pool them with a permutation-invariant readout before the head.
Our study is organized around a two-dimensional design space: three diffusion mechanisms that propagate information and three restriction-map parameterizations (and dimensions) that transport features between stalks. The two axes are orthogonal, so their combination forms a $3\times 3$ grid, one row of which is a new sheaf-attention layer we introduce. We first define the grid, then propose a Laplacian-free implementation for any diffusion mechanism that makes it trainable under cross-graph batching, introduce our dynamic sheaf-attention mechanism, derive the complexity, and define how modern GNN components are combined with the inductive sheaf operator (Fig.~\ref{fig:main-figure}-Top-left).

\paragraph{The Sheaf Design Space.}
\label{paragraph:design-space}

The three diffusion mechanisms analyzed in this study share the sheaf transport block ${P}_{ij}=\mathcal{F}_{i\trianglelefteq e}^{\top}\mathcal{F}_{j\trianglelefteq e}$, the mixing matrices ${W}_1\in\mathbb{R}^{d\times d}$ and ${W}_2\in\mathbb{R}^{f\times f}$, the intrinsic nonlinearity $\sigma$, and a residual connection. They differ in how neighbor messages are weighted. \textsc{NSD} propagates through the learnable normalized sheaf Laplacian of Eq.~\eqref{eq:nsd}, so every neighbor contributes through the transport alone. \textsc{SAN} and our newly proposed \textsc{SANv2} instead modulate the transport with input-dependent attention, and differ from each other in how the attention scores are computed: statically for \textsc{SAN}, dynamically for \textsc{SANv2}. Because the restriction maps set ${P}_{ij}$ while the attention coefficients set the weighting, \emph{how} features are transported between stalks stays decoupled from \emph{how strongly} each neighbor contributes.
Each mechanism is paired with one of the three restriction-map families seen in Sec.~\ref{sec:preliminaries} (\textsc{Diag}, \textsc{O(d)}, \textsc{Gen}), giving nine models. With all else fixed, this grid isolates two questions prior work conflates: (\textbf{1}) whether restriction-map type or diffusion mechanism dominates, and (\textbf{2}) whether they interact, that is, whether the best map family depends on the mechanism and the reverse.

\paragraph{Laplacian-Free Inductive Implementation.}
\label{paragraph:implementation}

The reference \textsc{NSD} implementation \citep{bodnar2022neural} targets a single fixed graph and cannot batch across graphs without restructuring. We instead implement every grid cell as a message-passing operator that never materializes the heavy $nd\times nd$ sheaf Laplacian and evaluates its action edge-wise, with the restriction maps $\mathcal{F}_{i\trianglelefteq e}$ predicted from \texttt{edge\_index} and its  (Fig.~\ref{fig:main-figure}-Top-right). For each edge $\{i,j\}\in E$ with edge features $\bm{e}_{ij}$, each layer computes three steps:
\begin{align}
  \text{(message)}\quad & {m}_{ij}=\phi(\bm{x}_i,\bm{x}_j,\bm{e}_{ij}); \label{eq:mp-message}\\
  \text{(aggregation)}\quad & \bm{a}_i=\bigoplus_{j\in\mathcal{N}(i)}{m}_{ij}
    =\sum_{j\in\mathcal{N}(i)}{m}_{ij}; \label{eq:mp-agg}\\
  \text{(update)}\quad & \bm{x}'_i=\gamma(\bm{x}_i,\bm{a}_i). \label{eq:mp-update}
\end{align}
In every case the message $\phi$ routes the neighbor's representation through the per-edge transport ${P}_{ij}=\mathcal{F}_{i\trianglelefteq e}^{\top}\mathcal{F}_{j\trianglelefteq e}$, 
while an optional mechanism-specific edge gate ${\Gamma}_{ij}$ 
and an optional node-wise normalization determine the exact form of $\phi$ and of the update $\gamma$. 

\textsc{NSD} applies no attention mechanism, ${\Gamma}_{ij}={I}_f$, and normalizes each stalk by its block degree
$
  {D}_i=\sum_{e\in\mathcal{E}(i)}\mathcal{F}_{i\trianglelefteq e}^{\top}\mathcal{F}_{i\trianglelefteq e},
$
where $\mathcal{E}(i)$ is the set of edges incident to $i$. Writing $\bm{z}_i={D}_i^{-1/2}{\bm{x}}_i$ for the normalized representation, the aggregation of Eq.~\eqref{eq:mp-agg} becomes the transported \emph{difference} between a node and its neighbors,
$
  \bm{a}_i={D}_i\bm{z}_i
    -\sum_{j\in\mathcal{N}(i)}{P}_{ij}\,\bm{z}_j,
$
and the update function, Eq.~\eqref{eq:mp-update}, is defined as follows:
\begin{equation}
  \gamma(\bm{x}_i,\bm{a}_i)=\bm{x}_i-\sigma\!\left({D}_i^{-1/2}\bm{a}_i\right).
\end{equation}
For the \textsc{Gen} parameterization we obtain ${D}_i^{-1/2}$ through a batched, deterministic Newton--Schulz iteration rather than the reference's random-noise regularization, which would otherwise be a source of numerical instability.

\textsc{SAN} and \textsc{SANv2} use the mixed representation $\tilde{\bm{x}}={W}_1\bm{x}{W}_2^{\top}$ directly, without degree normalization, and set ${\Gamma}_{ij}=\mathrm{diag}(\bm{\alpha}_{ij})$ with an attention gate $\bm{\alpha}_{ij}\in\mathbb{R}^{f}$, the channel-wise counterpart of the scalar coefficient $\Lambda_{ij}$ of Eq.~\eqref{eq:sheaf-attention}:
\begin{equation}
\label{eq:mp-san}
\begin{aligned}
  \bm{a}_i&=\sum_{j\in\mathcal{N}(i)}{P}_{ij}\,\tilde{\bm{x}}_j\,\mathrm{diag}\left(\bm{\alpha}_{ij} \right), \\
  \bm{x}'_i&=\bm{x}_i+\sigma\!\left(\bm{a}_i-\tilde{\bm{x}}_i\right),
\end{aligned}
\end{equation}
where $\tilde{\bm{x}}_i$ makes the bracket a high-pass sheaf update \citep{di2022understanding} distinct from the outer residual skip $\bm{x}_i$. 
The weights are produced by multi-head attention over per-edge scores $\alpha_{ij}$. \textsc{SAN} computes them with the static scoring of GAT \citep{velivckovic2018graph},
\begin{equation}
  \alpha_{ij}=\mathrm{LeakyReLU}\left(\bm{a}_s^{\top}{W}\bm{x}_i
      + \bm{a}_d^{\top}{W}\bm{x}_j\right).
  \label{eq:san}
\end{equation}
This scoring is \emph{static}: the ranking a node induces over its neighbors is fixed by the learned parameters and does not depend on the query node, a limitation that \citet{brody2021attentive} identified for GAT and resolved with GATv2. We therefore introduce \textsc{SANv2}, which keeps the sheaf update of Eq.~\eqref{eq:mp-san} unchanged, same transport ${P}_{ij}$, same channel-wise gate, same high-pass self term, and replaces only the scoring with its dynamic counterpart,
\begin{equation}
  \alpha_{ij}=\bm{a}^{\top}\mathrm{LeakyReLU}\left({W}_l\bm{x}_i
      + {W}_r\bm{x}_j\right).
  \label{eq:sanv2}
\end{equation}
By applying the nonlinearity after the combined projection and moving the attention vector outside it, GATv2 reordering makes neighbor ranking query-dependent rather than fixed \citep{brody2021attentive}. The change is confined to how each edge is scored, so it leaves the sheaf geometry, the complexity, and the batching properties of the layer untouched, and it composes with all three restriction-map families. To our knowledge \textsc{SANv2} is the first sheaf-attention layer with dynamic attention.
Scoring is multi-head with $H$ heads (head-specific parameters in Eqs.~\eqref{eq:san}--\eqref{eq:sanv2}), giving per-head coefficients $\alpha_{ij}^{(h)}=\mathrm{softmax}_j(e_{ij}^{(h)})$. By default the heads are combined by concatenation: head $h$ weights its own contiguous block of the $f$ channels, so $\bm{\alpha}_{ij}$ is block-constant across channels ($[\bm{\alpha}_{ij}]_c=\alpha_{ij}^{h(c)}$ for the head $h(c)$ owning channel $c$), while the transport ${P}_{ij}$ stays a single $d\times d$ block per edge.


\paragraph{Complexity.}
\label{paragraph:complexity}
Per layer per graph, restriction-map prediction dominates at
$\mathcal{O}(m\,d^3 f)$ for \textsc{O($d$)}/\textsc{Gen} and
$\mathcal{O}(m\,d^2 f)$ for \textsc{Diag}; transport and aggregation add $\mathcal{O}(m\,d^2 f)$; degree blocks and their inverse square roots add $\mathcal{O}((n+m)\,d^3)$. In total,
\begin{equation}
\label{eq:complexity}
\begin{aligned}
\text{Time:}\;& \mathcal{O}\left(\left(n+m \right)\,d^2 f + m\,d^3 f\right) \\
\text{Space:}\;& \mathcal{O}\big(m\,d^2 + (n+m)\,d f\big),
\end{aligned}
\end{equation}
both linear in $n$ and $m$. No operator is assembled or factorized, and the edge-wise pass composes with \textsc{PyTorch Geometric} batching, so a batch of graphs is one disjoint union processed without a per-graph loop. This is what makes our sheaf models trainable in the inductive, graph-batched regime the reference implementation does not support.

\paragraph{Composition with Modern GNN Components.}
\label{paragraph:gnnplus}

Two placements are specific to the sheaf operator. The activation $\sigma$ is intrinsic to the sheaf operator (Eqs.~\eqref{eq:nsd} and \eqref{eq:mp-san}), so no post-convolution activation is applied, and normalization and dropout act on the convolution increment \emph{before} the residual add. With $\mathrm{Nz}$ a normalization and $\mathrm{Dp}$ a dropout (both reduced to the identity when their component is switched off, i.e.\ $g_{\text{norm}}{=}0$ or $g_{\text{drop}}{=}0$), $g_\bullet\in\{0,1\}$ the component selectors, and input encodings concatenated upstream when $g_{\text{pe}}{=}1$, layer $l$ maps
\begin{align}
  \bm{a}^{(l)} &= \mathrm{Conv}_l\!\left(\bm{h}^{(l)}, E,
    g_{\text{edge}}\!\cdot\!\bm{e}_{ij}\right), \\
  \bm{b}^{(l)} &= \mathrm{Dp}\!\left(\mathrm{Nz}\left(\bm{a}^{(l)}\right)\right), \\
  \bm{c}^{(l)} &= g_{\text{res}}\,\bm{h}^{(l)} + \bm{b}^{(l)}, \\
  \bm{h}^{(l+1)} &= g_{\text{ffn}}\,\mathrm{FFN}_l\left(\bm{c}^{(l)}\right)
    + \left(1-g_{\text{ffn}} \right)\,\bm{c}^{(l)},
  \label{eq:gnnplus}
\end{align}
with the feed-forward block
\begin{equation*}  
\mathrm{FFN}_l(\bm{z})=\mathrm{Nz}_2\!\left(\mathrm{Nz}_1\left(\bm{z})
    +\mathrm{Dp}\left({W}_2^{\text{ff}}\,
    \mathrm{Dp}\left(\sigma({W}_1^{\text{ff}}\,\mathrm{Nz}_1(\bm{z})\right)\right)\right)\right)
  \label{eq:ffn}
\end{equation*}
The residual in Eq.~\eqref{eq:gnnplus} is the simple additive skip used by GNN$^{+}$. 
Edge features are integrated inside $\mathrm{Conv}_l$ by conditioning the
restriction maps rather than the messages or attention coefficients. The sheaf
learner predicts each map from the concatenated endpoint representations,
optionally augmented by an encoded edge feature,
\begin{equation*}
  \mathcal{F}^{(l)}_{v\trianglelefteq e}=\Phi_l\!\left(\left[\,\bm{h}^{(l)}_i
    \,\Vert\, \bm{h}^{(l)}_j \,\Vert\, g_{\text{edge}}\!\cdot\!
    {W}_e\,\bm{e}_{ij}\,\right]\right),
  \label{eq:edge-maps}
\end{equation*}
where $\Phi_l$ denotes the map-predicting linear projection followed by the family-specific construction (\textsc{DIAG}, $O(d)$, or \textsc{Gen}), and ${W}_e$ a learned edge encoder. When $g_{\text{edge}} =1$ the transport blocks $\mathcal{F}_i^{\top}\mathcal{F}_j$ and the degree normalization are thus conditioned on $\bm{e}_{ij}$. When $g_{\text{edge}}=0$ the maps depend on the node features alone as in the reference implementation. The selectors $g_\bullet$ parametrize the component configurations outlined in the following section.

\section{Experimental Setup}
\label{sec:experimental-setup}

\paragraph{Datasets and Tasks.} \paragraph{Datasets and Tasks.} We evaluate sheaf-based models on \emph{14} inductive datasets spanning graph- and node-level classification and regression tasks (Fig.~\ref{fig:main-figure}-Bottom). 
The evaluation includes five datasets from the GNN Benchmark~\citep{dwivedi2023benchmarking} (ZINC, MNIST, CIFAR10, PATTERN, and CLUSTER), four from the Open Graph Benchmark (OGB)~\citep{hu2020open} (ogbg-molhiv, ogbg-molpcba, ogbg-code2, and ogbg-ppa), and five from the Long-Range Graph Benchmark (LRGB)~\citep{dwivedi2022long, freitas2021large} (Peptides-func, Peptides-struct, PascalVOC-SP, COCO-SP, and MalNet-Tiny).
All 14 datasets are used in the screening stage, while the confirmation stage covers 12 for conciseness. We report the two remaining datasets, ogbg-ppa and COCO-SP, in the Supplementary Material.
We follow the official train/validation/test splits and evaluation protocols of each benchmark (Supplementary Material for details).
We follow the official train/validation/test splits and evaluation protocols of each benchmark (Supplementary Material for details).

\paragraph{Evaluation Protocol.} 
Following~\citet{luo2025can}, we adopt their standardized GNN$^{+}$ training pipeline\footnote{GNN$^{+}$  augments classic GNNs with normalization, residual connections, dropout, feed-forward layers, edge features, and positional/structural encodings~\cite{luo2025can}.}, including the same feature and positional/structural encoders, optimizer, learning-rate schedule, gradient clipping, and evaluation protocol (see Supplementary Material for details).
We replace only the message-passing layer with the proposed sheaf operator, ensuring that performance differences are attributable solely to the architecture.
%
We further adopt a two-stage \emph{screen-then-confirm} protocol. Candidate configurations are first screened using a single random seed and a reduced training budget, with model capacity constrained to the per-dataset parameter budget of~\citet{luo2025can}.
During the screening stage, we evaluate the two primary sheaf diffusion mechanisms, \textsc{NSD} and \textsc{SAN}, across the full factorial of datasets, restriction-map parameterizations, and stalk dimensions ($d \in \{2,3,4\}$), together with leave-one-out ablations of each GNN$^+$ component and an all-components-off baseline. 
Each component is evaluated across all restriction-map parameterizations and stalk dimensions, allowing conclusions to be drawn from aggregate trends over the full factorial design. For each diffusion mechanism and dataset, we select the best-performing configuration.
In the final confirmation stage, we evaluate \textsc{NSD}, \textsc{SAN}, and \textsc{SANv2}, where the latter inherits the best \textsc{SAN} configuration and is therefore not screened independently. All selected models are then trained using the full epoch and budget of Luo et al. 

\paragraph{Baselines.} 
Our comparisons differ between the two stages. During screening, leave-one-out ablations are evaluated against the corresponding all-off sheaf baseline to isolate the contribution of each GNN$^{+}$ component. During confirmation, the selected sheaf models are compared against strong GNNs equipped with the GNN$^{+}$ recipe, Graph Transformers, and prior specialized architectures, using the results reported in~\cite{luo2025can}. The \textcolor{green}{best}, \textcolor{blue}{second-}, and \textcolor{orange}{third-}best results are highlighted. We report the mean performance and standard deviation over five random seeds.

\section{Results}
\label{sec:results}

\subsection{Screening: Understanding the Sheaf Design Space}
\label{subsec:screening}
\begin{figure}
\centering
\includegraphics[width=\linewidth]{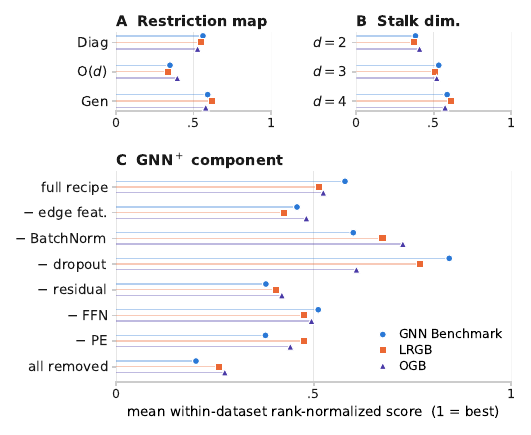}
\caption{Screening marginal effects. Each point is the mean within-dataset rank-normalized score ($1 =$ best cell on that dataset) at one level of a screening axis, averaged over the datasets of a benchmark suite. \textbf{A}: restriction-map parameterization. \textbf{B}: stalk dimension. \textbf{C}: GNN$^{+}$ component, where each row removes one component from the full recipe. Directions are consistent across all three suites. 
}
\label{fig:screening-marginals}
\end{figure}

The screening stage explores the sheaf design space across diffusion mechanisms, restriction-map parameterizations, stalk dimensions, and GNN$^{+}$ components to determine which design choices consistently govern the performance of inductive SNNs. Of its $14 \times 6 \times 3 \times 8 = 2{,}016$ cells, $108$ collapse onto the full recipe because six datasets carry no positional encoding and $18$ are infeasible at the parameter budget, leaving $1{,}890$ runs. Because the 14 datasets carry six metrics with opposite optimization directions, per-cell values cannot be pooled directly: we map each run to a within-dataset rank-normalized score in $[0,1]$ (Fig.~\ref{fig:screening-marginals}) and treat the \emph{dataset} as the unit of analysis, assessing every axis by paired comparisons across the 14 datasets with exact Wilcoxon signed-rank tests (Table~\ref{tab:ablation-main-effects}). Conclusions therefore hold across benchmarks rather than within individual cells. Since ranking discards the size of the gaps it orders, we repeated every comparison under a metric-linear score that retains them: $12$ of the $13$ are unchanged in direction and significance.

\begin{table}[htb!]
\centering
\resizebox{\columnwidth}{!}{
\begin{tabular}{lccc}
 \toprule
 Comparison & Mean $\Delta$ & Datasets favoring & $p$ \\
 \midrule
 \multicolumn{4}{l}{\emph{Restriction map} \quad ($\eta^2 = 12.2\%$; Friedman $\chi^2_2 = 17.71$, $p<0.001$)} \\
 \textsc{Gen} $>$ \textsc{O($d$)}   & $+0.237$ & 13/14 & $0.0004$ \\
 \textsc{Diag} $>$ \textsc{O($d$)}  & $+0.186$ & 13/14 & $0.0002$ \\
 \textsc{Gen} $>$ \textsc{Diag}     & $+0.052$ & 11/14 & $0.0052$ \\
 \midrule
 \multicolumn{4}{l}{\emph{Stalk dimension} \quad ($\eta^2 = 8.5\%$; Friedman $\chi^2_2 = 20.57$, $p<0.001$)} \\
 $d{=}4 > d{=}2$ & $+0.204$ & 13/14 & $0.0002$ \\
 $d{=}3 > d{=}2$ & $+0.132$ & 13/14 & $0.0002$ \\
 $d{=}4 > d{=}3$ & $+0.071$ & 13/14 & $0.0017$ \\
 \midrule
 \multicolumn{4}{l}{\emph{Diffusion mechanism} \quad ($\eta^2 = 0.2\%$)} \\
 \textsc{NSD} vs \textsc{SAN} & $+0.023$ & 8/14 & $0.502$ \\
 \midrule
 \multicolumn{4}{l}{\emph{GNN$^{+}$ component, removed vs.\ full recipe} \quad ($\eta^2 = 26.7\%$)} \\
 $-$ dropout   & $+0.211$ & 14/14 & $0.0001$ \\
 $-$ BatchNorm$^{\dagger}$ & $+0.124$ & 12/14 & $0.0085$ \\
 $-$ FFN       & $-0.045$ & \phantom{0}5/14 & $0.268$ \\
 $-$ PE        & $-0.109$ & \phantom{0}2/8\phantom{0}  & $0.148$ \\
 $-$ edge feat.& $-0.087$ & \phantom{0}2/14 & $0.0005$ \\
 $-$ residual  & $-0.139$ & \phantom{0}0/14 & $0.0001$ \\
 all removed   & $-0.294$ & \phantom{0}1/14 & $0.0006$ \\
 \bottomrule
\end{tabular}
}
\caption{Paired comparisons over 1,890 completed screening runs. $\Delta$ is the difference in within-dataset rank-normalized score, averaged over 12 datasets; $p$ is an exact two-sided Wilcoxon signed-rank test on paired per-dataset differences. $\eta^2$ is the share of total score variance attributable to the axis. $^{\dagger}$The only row whose significance depends on normalization: under a score linear in the native metric it falls to $p=0.058$, so we treat it as suggestive.}
\label{tab:ablation-main-effects}
\end{table}

\begin{table}[htb!]
\centering
\resizebox{\linewidth}{!}{
\begin{tabular}{l|ccccc}
 \toprule
 & ZINC & MNIST & CIFAR10 & PATTERN & CLUSTER \\
 \bottomrule
 Param. budget & $\sim500$K & $\sim100$K & $\sim100$K & $\sim500$K & $\sim500$K \\
 Metric & MAE$\downarrow$ & Accuracy$\uparrow$ & Accuracy$\uparrow$ & Accuracy$\uparrow$ & Accuracy$\uparrow$ \\
 \midrule
 EGT (\citeyear{hussain2022global}) & 0.108$\pm$0.009 & 0.9817$\pm$0.0009 & 0.6870$\pm$0.0041 & 0.8682$\pm$0.0002 & \textcolor{blue}{0.7923$\pm$0.0035} \\
 Graph ViT/MLP-Mixer (\citeyear{he2023generalization}) & 0.073$\pm$0.001 & \textcolor{orange}{0.9846$\pm$0.0009} & 0.7396$\pm$0.0033 & – & – \\
 GRIT (\citeyear{ma2023graph}) & \textcolor{blue}{0.059$\pm$0.002} & 0.9811$\pm$0.0011 & \textcolor{orange}{0.7647$\pm$0.0088} & \textcolor{green}{0.8720$\pm$0.0008} & \textcolor{green}{0.8003$\pm$0.0028} \\
 GRED (\citeyear{ding2024recurrent}) & 0.077$\pm$0.002 & 0.9838$\pm$0.0001 & \textcolor{blue}{0.7685$\pm$0.0019} & 0.8676$\pm$0.0002 & 0.7850$\pm$0.0010 \\
 GEAET (\citeyear{liang2024graph}) & – & \textcolor{blue}{0.9851$\pm$0.0009} & 0.7663$\pm$0.0043 & 0.8699$\pm$0.0003 & – \\
 TIGT (\citeyear{choi2024topology}) & \textcolor{green}{0.057$\pm$0.002} & 0.9823$\pm$0.0013 & 0.7396$\pm$0.0036 & 0.8668$\pm$0.0006 & 0.7802$\pm$0.0022 \\
 GMN (\citeyear{behrouz2024graph}) & – & 0.9839$\pm$0.0018 & 0.7456$\pm$0.0038 & \textcolor{blue}{0.8709$\pm$0.0126} & – \\
 GIN$^+$ (\citeyear{luo2025can}) & \textcolor{orange}{0.065$\pm$0.004} & 0.9829$\pm$0.0010 & 0.6959$\pm$0.0029 & 0.8684$\pm$0.0005 & 0.7479$\pm$0.0021 \\
 GatedGCN$^+$ (\citeyear{luo2025can}) & 0.077$\pm$0.005 & \textcolor{green}{0.9871$\pm$0.0014} & \textcolor{green}{0.7722$\pm$0.0038} & \textcolor{orange}{0.8703$\pm$0.0004} & \textcolor{orange}{0.7913$\pm$0.0024} \\
 \midrule
 \textsc{NSD} & 0.089$\pm$0.008 & 0.9751$\pm$0.0037 & 0.6661$\pm$0.0051 & 0.8674$\pm$0.0001 & 0.7836$\pm$0.0071 \\
 \textsc{SAN} & 0.088$\pm$0.003 & 0.9753$\pm$0.0012 & 0.6636$\pm$0.0059 & 0.8677$\pm$0.0003 & 0.7666$\pm$0.0041 \\
 \textsc{SANv2} & 0.092$\pm$0.004 & 0.9742$\pm$0.0015 & 0.6750$\pm$0.0063 & 0.8684$\pm$0.0002 & 0.6805$\pm$0.0678$^{\dagger}$ \\
 \bottomrule
\end{tabular}
}
\caption{Performance on GNN Benchmark datasets.}
\label{tab:GNN-benchmark-results}
\end{table}

\begin{table}[htb!]
\centering
\resizebox{\linewidth}{!}{
\begin{tabular}{l|cccc}
 \toprule
 & Peptides-func & Peptides-struct & PascalVOC-SP & MalNet-Tiny \\
 \bottomrule
 Param. budget & $\sim500$K & $\sim500$K & $\sim500$K & $\sim500$K \\
 Metric & Avg. Precision$\uparrow$ & MAE$\downarrow$ & F1 score$\uparrow$ & Accuracy$\uparrow$ \\
 \midrule
GraphGPS (\citeyear{rampavsek2022recipe}) & 0.6534$\pm$0.0091 & 0.2509$\pm$0.0014 & \textcolor{blue}{0.4440$\pm$0.0065} & 0.9350$\pm$0.0041 \\
Exphormer (\citeyear{shirzad2023exphormer}) & 0.6258$\pm$0.0092 & 0.2512$\pm$0.0025 & 0.3446$\pm$0.0064 & \textcolor{orange}{0.9402$\pm$0.002} \\
GRED (\citeyear{ding2024recurrent}) & \textcolor{blue}{0.7133$\pm$0.0011} & 0.2455$\pm$0.0013 & – & – \\
Graph-Mamba (\citeyear{wang2024graph}) & 0.6739$\pm$0.0087 & 0.2478$\pm$0.0016 & 0.4191$\pm$0.0126 & 0.9340$\pm$0.0027 \\
GSSC (\citeyear{huang2024can}) & \textcolor{orange}{0.7081$\pm$0.0062} & 0.2459$\pm$0.0020 & \textcolor{green}{0.4561$\pm$0.0039} & \textcolor{blue}{0.9406$\pm$0.0064} \\
GCN+ (\citeyear{luo2025can}) & \textcolor{green}{0.7261$\pm$0.0067} & \textcolor{green}{0.2421$\pm$0.0016} & 0.3357$\pm$0.0087 & 0.9354$\pm$0.0045 \\
GIN+ (\citeyear{luo2025can}) & 0.7059$\pm$0.0089 & \textcolor{blue}{0.2429$\pm$0.0019} & 0.3189$\pm$0.0105 & 0.9325$\pm$0.0040 \\
GatedGCN+ (\citeyear{luo2025can}) & 0.7006$\pm$0.0033 & \textcolor{orange}{0.2431$\pm$0.0020} & \textcolor{orange}{0.4263$\pm$0.0057} & \textcolor{green}{0.9460$\pm$0.0057} \\
 \midrule
 \textsc{NSD} & 0.6374$\pm$0.0071 & 0.2577$\pm$0.0021 & 0.3317$\pm$0.0069 & 0.9300$\pm$0.0014 \\
 \textsc{SAN} & 0.6376$\pm$0.0075 & {0.2583$\pm$0.0021} & 0.3251$\pm$0.0035 & 0.9235$\pm$0.0035 \\
 \textsc{SANv2} & 0.6228$\pm$0.0093 & 0.2546$\pm$0.0037 & 0.3052$\pm$0.0008 & 0.9275$\pm$0.0078 \\
 \bottomrule
\end{tabular}
}
\caption{Performance on LRGB datasets.}
\label{tab:LRGB-benchmark-results}
\end{table}

\begin{table}[htb!]
\centering
\resizebox{\linewidth}{!}{
\begin{tabular}{l|ccc}
 \toprule
 & ogbg-molhiv & ogbg-molpcba & ogbg-code2 \\
 \bottomrule
 Metric & AUROC$\uparrow$ & Avg. Precision$\uparrow$ & F1 score$\uparrow$ \\
 \midrule
SAT (\citeyear{chen2022structure}) & – & – & \textcolor{green}{0.1937 ± 0.0028} \\
Specformer (\citeyear{bo2023specformer}) & 0.7889 ± 0.0124 & \textcolor{blue}{0.2972 ± 0.0023} & – \\
Subgraphormer (\citeyear{bar2024subgraphormer}) & \textcolor{blue}{0.8038 ± 0.0192} & – & – \\
GECO (\citeyear{sancak2024scalable}) & 0.7980 ± 0.0200 & \textcolor{orange}{0.2961 ± 0.0008} & \textcolor{blue}{0.1915 ± 0.0020} \\
GSSC (\citeyear{huang2024can}) & \textcolor{orange}{0.8035 ± 0.0142} & – & – \\
GCN+ (\citeyear{luo2025can}) & 0.8012 ± 0.0124 & 0.2721 ± 0.0046 & 0.1787 ± 0.0026 \\
GIN+ (\citeyear{luo2025can}) & 0.7928 ± 0.0099 & 0.2703 ± 0.0024 & 0.1803 ± 0.0019 \\
GatedGCN+ (\citeyear{luo2025can}) & \textcolor{green}{0.8040 ± 0.0164} & \textcolor{green}{0.2981 ± 0.0024} & \textcolor{orange}{0.1896 ± 0.0024} \\
 \midrule
 \textsc{NSD} & 0.7606$\pm$0.0021 & 0.2333$\pm$0.0014 & 0.1559$\pm$0.0197 \\
 \textsc{SAN} & 0.7563$\pm$0.0275 & 0.2349$\pm$0.0031 & 0.1592$\pm$0.0219 \\
 \textsc{SANv2} & 0.7241$\pm$0.0202 & 0.2362$\pm$0.0008 & 0.1594$\pm$0.0205 \\
 \bottomrule
\end{tabular}
}
\caption{Performance on OGB datasets.}
\label{tab:OGB-benchmark-results}
\end{table}

Several insights are revealed. \textbf{Restriction-map parameterization is the dominant design choice}, accounting for $12.2\%$ of the score variance against $0.2\%$ for the choice of diffusion mechanism, with the ordering $\textsc{Gen} > \textsc{Diag} > \textsc{O}(d)$ holding in all six mechanism~$\times$~$d$ slices. The \textsc{Gen}--\textsc{O}($d$) gap is $0.242$, $0.244$ and $0.224$ at $d{=}\{2,3,4\}$, hence independent of how many free parameters an orthogonal map is given. This suggests that \textbf{orthogonal transport is limited by its norm-preserving constraint rather than by its smaller parameterization}, since preserving feature norms limits the ability to adaptively attenuate neighboring messages. Consistently, \textsc{O}($d$) fares worst where edges are most numerous ($0.316$ on the four superpixel datasets against $0.395$ on OGB).
\textbf{Increasing the stalk dimension improves performance monotonically on $13/14$ datasets}, with every successive step significant (except ogbg-ppa, discussed below). Setting larger $d$, each node receives extra channel capacity to route different neighbors' contributions. Such added capacity is what sheaf diffusion is argued to gain over scalar-weighted message passing. However, if observed gains came from that mechanism, they should be concentrated where information must travel farthest, or where many neighbors compete for the same channels, but \emph{we observe no such concentration}. Over the $13$ datasets on which the trend is monotone, the $d{=}4$ over $d{=}2$ improvement is uncorrelated with the average number of nodes per graph (Spearman $\rho = +0.08$) and with the average node degree ($\rho = -0.15$), and its mean over the five LRGB datasets ($+0.234$), which are constructed to require long-range dependencies, is close to its mean over the remaining eight ($+0.218$). Therefore, \textbf{larger stalks behave as a generic increase in model capacity rather than as a targeted remedy for long-range reach}, although with $13$ datasets only a strong association would be detectable.

\textbf{Searching the sheaf design space per dataset recovers almost nothing once the GNN$^{+}$ recipe is tuned.} Selecting the best of the $18$ architectures separately on every dataset attains a rank-normalized score of $1.0$ there by construction, yet fixing the architecture globally to \textsc{NSD}-\textsc{Gen} at $d{=}4$ and tuning only the recipe per dataset still averages $0.991$, and is the single best cell outright on five of the $14$ datasets. A per-dataset tuning budget is therefore better spent on the GNN$^{+}$ components than on the sheaf architecture.

\textbf{The surrounding GNN$^{+}$ architecture explains more performance variation ($26.7\%$) than the entire 18-cell sheaf design space ($21.1\%$)}. In particular, \textbf{residual connections and edge features are essential}, with only $0/14$ and $2/14$ datasets improving when they are removed, whereas \textbf{dropout is detrimental on all $14$ datasets}. A plausible explanation is that the learned restriction maps already act as a per-edge bottleneck, providing an implicit form of regularization that reduces the benefit of dropout.

Finally, while \textsc{NSD} and \textsc{SAN} achieve comparable average performance ($p = 0.50$), they exhibit different behaviors: \textbf{\textsc{SAN} is more robust to the removal of GNN$^{+}$ components}, losing $0.169$ score points against $0.422$ for \textsc{NSD} when all are removed, and it leads on $5/8$ graph-level datasets, whereas \textbf{\textsc{NSD} is ahead on all four node-level datasets} (mean $+0.126$). However, the two mechanisms do not merely degrade at different rates: \textbf{their ranking reverses with the component stack}. Under the full recipe \textsc{NSD} leads, $0.564$ against $0.515$; with every component removed \textsc{SAN} leads by a wider margin, $0.346$ against $0.142$. The aggregate parity reported above is thus an average over two opposite regimes, and a comparison between diffusion mechanisms is interpretable only with respect to the component stack in which it was measured.

One dataset, {ogbg-ppa}, deviates from the overall trends. Under every recipe retaining \texttt{BatchNorm}, as prescribed by the evaluation protocol, the models fit the training set while validation accuracy collapses to at most $11\%$. Removing \texttt{BatchNorm} leaves training essentially unchanged but restores validation performance (see Supplementary Material), suggesting that estimated batch statistics do not transfer to evaluation. This sensitivity appears specific to sheaf diffusion rather than to the dataset itself, as GNN$^{+}$ baselines achieve $0.81$--$0.83$ under the same protocol. We therefore retain the reference normalization for a fair comparison and simply note that default ranking compares near-chance models. This also highlights a constraint of the inherited protocol that prevents from using alternative normalizations that behave identically during training and evaluation (e.g. \texttt{LayerNorm}). While we cannot relax it, we leave this exploration to future studies.
Overall, these findings identify expressive restriction maps, larger stalk dimensions, and a carefully designed GNN$^{+}$ architecture as the primary factors behind SNN's inductive performance. Performance of individual cells is provided in the Supplementary Material.

\subsection{Confirmation: Overall Performance}

Following the screening stage, Tables~\ref{tab:GNN-benchmark-results}--\ref{tab:OGB-benchmark-results} report the selected \textsc{NSD}, \textsc{SAN}, and \textsc{SANv2} configurations alongside strong GNNs, GTs, and specialized baselines under the standardized evaluation protocol of~\citet{luo2025can}. Performance is reported as the mean and standard deviation over five random seeds. Only top-3 entries from the work of \citet{luo2025can} are reported for conciseness.

Overall, SNNs can transfer to inductive settings, although a gap with the strongest modern architectures remains. That gap is highly dataset dependent, spanning two orders of magnitude across the twelve datasets reported here, and it tracks neither the benchmark suite nor the task level: the smallest and largest gaps both occur within the GNN Benchmark suite, and node- and graph-level datasets are interleaved throughout the ranking.
The closest results are obtained on PATTERN, MNIST, MalNet-Tiny and CLUSTER, where sheaf models average $1.4\%$ from the best-performing method, with \textsc{NSD} outperforming TIGT and GIN$^{+}$ on CLUSTER. The largest gaps, averaging $30\%$, are on ZINC, PascalVOC-SP, ogbg-molpcba and ogbg-code2. Notably, the four datasets on which sheaf models come closest are also those on which the baselines themselves differ least, so near-parity there suggests limited headroom rather than a property of sheaf diffusion.
Comparing the three mechanisms reveals complementary behavior rather than a universally superior architecture. \textsc{NSD} attains the best sheaf result on two of the three node-classification datasets, in line with transductive findings~\citep{barbero2022sheaf, bodnar2021weisfeiler}, whereas the attention-based mechanisms take seven of the nine graph-level datasets. Although \textsc{SANv2} introduces dynamic GATv2-style attention, it does not consistently outperform the original \textsc{SAN}, suggesting that attention design alone is not the primary bottleneck for inductive sheaf learning.
Overall, these observations closely mirror the conclusions of the screening study. Restriction-map parameterization and the surrounding GNN$^{+}$ architecture account for substantially more performance variation than the choice of diffusion mechanism, while \textsc{NSD} and \textsc{SAN} exhibit complementary inductive biases that persist under full training. These results indicate that the principal limitations of current SNNs lie less in the diffusion mechanism itself than in the architectural design surrounding it, providing a clear direction for future sheaf-based models.

\section{Conclusions}
\label{sec:conclusions}
We explored the sheaf design space across diffusion mechanisms, restriction-map parameterizations, stalk dimensions, and modern GNN components on 14 inductive graph- and node-level datasets under a standardized protocol, screening $1{,}890$ configurations and confirming the best ones at full budget. SNNs can transfer to inductive settings, but do not reach the strongest published baselines on any dataset considered, and the gap varies by two orders of magnitude across datasets without tracking benchmark suite or task level. Screening identifies where performance is decided: restriction-map parameterizations govern the sheaf operator, \textsc{GEN} maps are preferable at every stalk dimension, larger stalks contribute capacity rather than long-range reach, and the surrounding component recipe explains more variance than the entire sheaf-specific design. Practically, a single sheaf architecture can generalize across datasets while the component recipe does not, so tuning is better directed at the latter. This suggests where future SNNs should invest: expressive maps and component recipe.
\textbf{Limitations and Future Works.} Despite our extensive experiments, our factorial exploration is only near-complete, and additional interactions among the GNN$^{+}$ components may provide further insights. Likewise, although we evaluate a broad collection of datasets, stronger correlations may emerge only from a larger and more diverse benchmark suite. We leave both directions for future work.

\clearpage

\bibliography{bibliography}


\clearpage

\section{Experimental Setup and Reproducibility}

\subsection{Code}
The code associated with this work will be released upon acceptance for fostering research on Sheaf Neural Networks.

\subsection{Computing Environment}

All experiments are run, one per GPU, on NVIDIA DGX A100 (with 8 GPUs of 80GB memory, optionally partitioned in MIG instances of 40GB for lighter experiments), a double 64-core AMD EPYC 7742 and 2TB of RAM. The
software environment is fully specified by the \texttt{environment.yml} file
shipped with the code and can be recreated with a single
\texttt{conda env create} invocation. The versions of the components that
determine numerical behavior are listed in Table~\ref{tab:supp-env}. We do not
use mixed precision, and all reported numbers are produced in float32.

\begin{table}[htb!]
\centering
\begin{tabular}{ll}
 \toprule
 Component & Version \\
 \midrule
 Python              & 3.10.20 \\
 PyTorch             & 2.2.0 (+cu118) \\
 CUDA toolkit        & 11.8 \\
 PyTorch Geometric   & 2.3.1 \\
 \texttt{pyg-lib}    & 0.4.0 \\
 \texttt{torch-scatter}  & 2.1.2 \\
 \texttt{torch-sparse}   & 0.6.18 \\
 \texttt{torch-cluster}  & 1.6.3 \\
 \texttt{torchmetrics}   & 1.9.0 \\
 GPU                 & NVIDIA A100 40--80\,GB \\
 \bottomrule
\end{tabular}
\caption{Software and hardware environment. The complete dependency set,
including transitive pins, is given in \texttt{environment.yml} in the released
code.}
\label{tab:supp-env}
\end{table}

\subsection{What Is Inherited and What Is Replaced}

Our implementation reuses, without modification, the GraphGym-based training and evaluation pipeline released by~\citet{luo2025can}. Only the message-passing block is replaced by the sheaf operator. Specifically, we inherit verbatim the per-dataset node and edge feature encoders, Random-Walk Structural Encoding (RWSE) where applicable, the AdamW optimizer, the cosine learning-rate schedule with warmup, gradient clipping, weight decay, batch sizes, loss functions, metric implementations, and the train/validation/test splits. Any difference in performance is therefore attributable to the architecture rather than to the training recipe, so that the results we report are directly comparable to those of~\citet{luo2025can}.

We acknowledge that this inheritance may also represent a constraint in some cases. For example, the normalization layer is exposed by the pipeline as a binary switch rather than as a choice of scheme, which is why the \texttt{BatchNorm} behavior on ogbg-ppa discussed in the main paper could be diagnosed but not altered within the protocol. 

\subsection{Seeding and Determinism}

Every run is seeded through a single entry point before any model, data loader, or shuffling is constructed, using \texttt{torch\_geometric.seed\_everything}, which sets the Python, NumPy, and PyTorch (CPU and CUDA) generators. A run with index $i$ uses seed $s_0+i$, with $s_0 = 0$ throughout, so the seeds of a $k$-seed confirmation are $0, \ldots, k-1$ and are identical across architectures and datasets. Seeds control weight initialization, data-loader shuffling, dropout masks, and any stochastic component of the restriction-map learner. Notably, screening uses a single - design (Section~\ref{app:screnconfirm}), since its purpose is to identify aggregate trends and consistent ranking across many configurations rather than to support claims about individual cells. Nonetheless, the insights reported in the main paper are supported by solid statistical tests and analysis. Confirmation stage adopts 5 seeds, consistently with prior literature.

\subsection{Screen-then-Confirm Protocol}
\label{app:screnconfirm}
The evaluation proceeds in two stages:
\begin{enumerate}
    \item \textbf{Screening.} A near-complete factorial over diffusion mechanism, restriction-map family, stalk dimension, and GNN$^{+}$ component ablations is run with a single seed and the reduced per-dataset epoch budget of Table~\ref{tab:supp-budget}. The purpose of this stage is to rank candidate configurations consistently and aggregate trends rather than to estimate their final performance.
    \item \textbf{Confirmation.} For each dataset and each diffusion mechanism, the configuration with the best validation score is retrained from scratch at the full reference epoch budget and matched per-dataset parameter budget of~\citet{luo2025can}, over five random seeds. SANv2 inherits the SAN configuration rather than being screened independently, so that the comparison between the two isolates the attention mechanism.
\end{enumerate}

\paragraph{Model selection.}
In both stages, the reported test metric is the one recorded at the epoch of best validation performance (test-at-best-validation). The test split is never used for model selection or hyperparameter tuning.


\begin{table}[htb!]
\centering
\begin{tabular}{lr}
 \toprule
 Dataset & Screen epochs \\
 \midrule
 ZINC            & 200 \\
 MNIST           & 20  \\
 CIFAR10         & 20  \\
 PATTERN         & 20  \\
 CLUSTER         & 10  \\
 Peptides-func   & 50  \\
 Peptides-struct & 30  \\
 PascalVOC-SP    & 20  \\
 COCO-SP         & 15  \\
 MalNet-Tiny     & 15  \\
 ogbg-molhiv     & 30  \\
 ogbg-molpcba    & 10  \\
 ogbg-ppa        & 10  \\
 ogbg-code2      & 5   \\
 \bottomrule
\end{tabular}
\caption{Reduced per-dataset epoch budget used in the screening stage. The confirmation stage uses the full reference budget of \citet{luo2025can}.}
\label{tab:supp-budget}
\end{table}
\subsection{Fair-Benchmarking Choices}

We list the decisions that most affect comparability, so that they can be audited rather than inferred.

\begin{itemize}

\item \textbf{Matched pipeline.} Every component outside the message-passing block is inherited from~\citet{luo2025can} and is identical across all architectures compared, including the baselines whose numbers we quote.

\item \textbf{Matched parameter budget.} Confirmation widths are chosen so that the parameter count matches as closely as possible the values reported by~\citet{luo2025can}. This ensures the fairest comparison across architectures.

\item \textbf{Official splits.} All datasets use their prescribed train/validation/test splits and evaluation metrics. No split is re-partitioned and no metric is redefined.

\item \textbf{No test-set selection.} Model selection is performed exclusively on the validation set in both screening and confirmation.

\item \textbf{Baselines are quoted, not re-run.} All baseline numbers are taken from the original publications. Comparability rests on the shared evaluation pipeline and the matched parameter budget rather than on re-training.

\item \textbf{Negative and anomalous results are reported.} For transparency, we report all completed experiments, including unsuccessful configurations and anomalous behaviors such as the \texttt{BatchNorm} sensitivity on ogbg-ppa and the infeasible SAN configurations on PascalVOC-SP (Figs~\ref{fig:supp-heat-gnn}-\ref{fig:supp-heat-ogb}).
\end{itemize}

\section{Datasets}

\subsection{Selection Rationale}

We evaluate on the same 14 inductive benchmarks used by \citet{luo2025can}.
We adopt an existing,
widely used suite in full and remove the possibility of favorable dataset
selection. This makes every number we report directly comparable to a
published set of strong baselines evaluated under an identical pipeline.
The suite spans:
graph- and node-level tasks, classification and regression, molecular, image,
synthetic, program, and biological graphs, and graph sizes from 23 to 1,410
nodes on average. 

\subsection{Descriptions}

\begin{itemize}
\item \textbf{GNN Benchmark} \citep{dwivedi2023benchmarking}. \textbf{ZINC}
contains molecular graphs (atom node features, bond edge features); the task is
to regress the constrained solubility (logP). \textbf{MNIST} and
\textbf{CIFAR10} are image-classification datasets recast as $8$-nearest-neighbor
graphs of SLIC superpixels, retaining the original $10$-class tasks.
\textbf{PATTERN} and \textbf{CLUSTER} are synthetic Stochastic Block Model graphs
for inductive node classification (pattern recognition and cluster-ID inference,
respectively).
\item \textbf{Long-Range Graph Benchmark (LRGB)} \citep{dwivedi2022long,
freitas2021large}. \textbf{Peptides-func} and \textbf{Peptides-struct} are atomic
peptide graphs, with $10$-way multi-label graph classification and $11$-target
structural regression. \textbf{PascalVOC-SP} and \textbf{COCO-SP} are superpixel
node-classification datasets derived from Pascal VOC and MS COCO.
\textbf{MalNet-Tiny} is a subset of function-call graphs from Android APKs, with
software-type classification from structure alone.
\item \textbf{Open Graph Benchmark (OGB)} \citep{hu2020open}.
\textbf{ogbg-molhiv} and \textbf{ogbg-molpcba} are molecular property prediction
datasets (binary HIV-inhibition classification and a $128$-task bioassay panel).
\textbf{ogbg-ppa} classifies protein-association networks into $37$ taxonomic
groups, and \textbf{ogbg-code2} predicts the first five subtokens of a
function's name from its abstract syntax tree.
\end{itemize}

\begin{table}[htb!]
\centering
\resizebox{\columnwidth}{!}{
\begin{tabular}{lrrrcc}
 \toprule
 Dataset & \# graphs & Avg. \# nodes & Avg. \# edges & Task type & Metric \\
 \midrule
 ZINC & 12,000 & 23.2 & 24.9 & Graph regr. & MAE \\
 MNIST & 70,000 & 70.6 & 564.5 & Graph cls. & Accuracy \\
 CIFAR10 & 60,000 & 117.6 & 941.1 & Graph cls. & Accuracy \\
 PATTERN & 14,000 & 118.9 & 3,039.3 & Node cls. & Accuracy \\
 CLUSTER & 12,000 & 117.2 & 2,150.9 & Node cls. & Accuracy \\
 \midrule
 Peptides-func & 15,535 & 150.9 & 307.3 & Graph cls. & Avg. Precision \\
 Peptides-struct & 15,535 & 150.9 & 307.3 & Graph regr. & MAE \\
 PascalVOC-SP & 11,355 & 479.4 & 2,710.5 & Node cls. & F1 score \\
 COCO-SP & 123,286 & 476.9 & 2,693.7 & Node cls. & F1 score \\
 MalNet-Tiny & 5,000 & 1,410.3 & 2,859.9 & Graph cls. & Accuracy \\
 \midrule
 ogbg-molhiv & 41,127 & 25.5 & 27.5 & Graph cls. & AUROC \\
 ogbg-molpcba & 437,929 & 26.0 & 28.1 & Graph cls. & Avg. Precision \\
 ogbg-ppa & 158,100 & 243.4 & 2,266.1 & Graph cls. & Accuracy \\
 ogbg-code2 & 452,741 & 125.2 & 124.2 & Graph cls. & F1 score \\
 \bottomrule
\end{tabular}
}
\caption{Statistics of the 14 inductive datasets used in this paper.}
\label{tab:supp-datasets}
\end{table}

\begin{table*}[htb!]
\centering
\resizebox{0.8\linewidth}{!}{
\begin{tabular}{lll}
 \toprule
 Resource & License & Source \\
 \midrule
 GNN Benchmark suite & MIT & \url{github.com/graphdeeplearning/benchmarking-gnns} \\
 \midrule
 LRGB release & CC BY 4.0 & \url{doi.org/10.5281/zenodo.6975830} \\
 \quad PascalVOC-SP & Custom (Flickr ToU) & Pascal VOC 2011 \\
 \quad COCO-SP & CC BY 4.0 & MS COCO \\
 \quad Peptides-func/struct & Public-domain sources & SATPdb \\
 \midrule
 MalNet-Tiny & CC BY 4.0 & \url{mal-net.org} \\
 \midrule
 OGB datasets & MIT & \url{ogb.stanford.edu} \\
 \midrule
 GNN$^{+}$ codebase & MIT & \url{github.com/LUOyk1999/GNNPlus} \\
 \bottomrule
\end{tabular}
}
\caption{Licenses of the datasets and code used in this work.}
\label{tab:supp-licenses}
\end{table*}

\begin{table}[htb!]
\centering
\resizebox{\columnwidth}{!}{
\begin{tabular}{lcc}
 \toprule
 & COCO-SP & ogbg-ppa \\
 Metric & F1 score $\uparrow$ & Accuracy $\uparrow$ \\
 \midrule
 Graph-Mamba & \textcolor{green}{0.3960 $\pm$ 0.0175} & - \\
 GraphGPS & \textcolor{blue}{0.3884 $\pm$ 0.0050} & - \\
 GCN$^{+}$ & 0.2733 $\pm$ 0.0041 & \textcolor{orange}{0.8077 $\pm$ 0.0041} \\
 GIN$^{+}$      & 0.2483 $\pm$ 0.0046 & \textcolor{blue}{0.8107 $\pm$ 0.0053} \\
 GatedGCN$^{+}$  & \textcolor{orange}{0.3802 $\pm$ 0.0015} & \textcolor{green}{0.8258 $\pm$ 0.0055} \\
 \midrule
 \textsc{NSD}   & 0.3004 $\pm$ 0.0027 & 0.6752 $\pm$ 0.0047 \\
 \textsc{SAN}   & 0.2979 $\pm$ 0.0034 & 0.6744 $\pm$ 0.0374 \\
 \textsc{SANv2} & 0.2971 $\pm$ 0.0029 & 0.6535 $\pm$ 0.0474 \\
 \bottomrule
\end{tabular}
}
\caption{Confirmation results on COCO-SP and ogbg-ppa. All results except the sheaf models are taken from \citet{luo2025can}}
\label{tab:supp-confirm-full}
\end{table}

\begin{table*}[htb!]
\centering
\resizebox{\linewidth}{!}{
\begin{tabular}{lcccrrcccrrcccrr}
 \toprule
 & \multicolumn{5}{c}{\textsc{NSD}} & \multicolumn{5}{c}{\textsc{SAN}} & \multicolumn{5}{c}{\textsc{SANv2}} \\
 Dataset & Map & $d$ & GNN$^+$ & $L$ & width  & Map & $d$ & GNN$^+$ & $L$ & width  & Map & $d$ & GNN$^+$ & $L$ & width \\
 \midrule
 ZINC            & diag & 4 & $-$drop & 9  & 96  & diag & 4 & $-$drop & 9  & 60  & diag & 4 & $-$drop & 9  & 40  \\
 MNIST           & gen  & 4 & $-$drop & 10 & 40  & diag & 4 & $-$drop & 10 & 40  & diag & 4 & $-$drop & 10 & 32  \\
 CIFAR10         & gen  & 4 & $-$drop & 10 & 40  & diag & 4 & $-$drop & 10 & 40  & diag & 4 & $-$drop & 10 & 36  \\
 PATTERN         & gen  & 3 & $-$drop & 12 & 81  & diag & 4 & $-$drop & 12 & 48  & diag & 4 & $-$drop & 12 & 32  \\
 CLUSTER         & gen  & 4 & $-$norm & 16 & 68  & gen  & 4 & $-$drop & 16 & 36  & gen  & 4 & $-$drop & 16 & 24  \\
 \midrule
 Peptides-func   & gen  & 3 & $-$drop & 5  & 120 & diag & 4 & $-$drop & 5  & 88  & diag & 4 & $-$drop & 5  & 64  \\
 Peptides-struct & gen  & 4 & $-$drop & 4  & 132 & gen  & 4 & $-$norm & 4  & 100 & gen  & 4 & $-$norm & 4  & 80  \\
 PascalVOC-SP    & gen  & 4 & $-$drop & 12 & 84  & diag & 4 & $-$drop & 12 & 52  & diag & 4 & $-$drop & 12 & 36  \\
 COCO-SP         & gen  & 4 & $-$drop & 20 & 60  & diag & 4 & $-$drop & 20 & 36  & diag & 4 & $-$drop & 20 & 20  \\
 MalNet-Tiny     & gen  & 4 & $-$drop & 6  & 120 & diag & 4 & $-$drop & 6  & 88  & diag & 4 & $-$drop & 6  & 64  \\
 \midrule
 ogbg-molhiv     & gen  & 3 & $-$res  & 3  & 255 & gen & 2 & $-$PE  & 3  & 256 & gen & 2 & $-$PE  & 3  & 256 \\
 ogbg-molpcba    & gen  & 4 & $-$drop & 10 & 256 & gen & 4 & $-$drop & 10 & 256 & gen & 4 & $-$drop & 10 & 256 \\
 ogbg-ppa        & gen  & 4 & $-$norm & 4  & 512 & gen & 4 & $-$norm & 4  & 512 & gen & 4 & $-$norm & 4  & 512 \\
 ogbg-code2      & gen  & 4 & $-$norm & 5  & 512 & gen & 4 & $-$norm & 5 & 512 & gen & 4 & $-$norm & 5 & 512 \\
 \bottomrule
\end{tabular}
}
\caption{Configurations promoted from screening to confirmation, selected per
dataset \emph{and} per diffusion mechanism on validation performance. ``Map'' is
the restriction-map parameterization, $d$ the stalk dimension, ``GNN$^+$'' is the
architectural component removal that ranked best ($-$drop $=$ dropout removed,
$-$norm $=$ BatchNorm removed, $-$res $=$ residual removed, $-$PE $=$ positional
encoding removed), $L$ the number of message-passing layers, and ``width'' the
hidden dimension. \textsc{SANv2} inherits \textsc{SAN}'s configuration rather than being screened independently.}
\label{tab:supp-confirm-configs}
\end{table*}

\subsection{Licenses and Availability}

Table~\ref{tab:supp-licenses} lists the license of every dataset and code resource used in this work. All datasets are publicly available and are obtained through their official benchmark loaders, and each remains subject to the license of its original provider. We do not redistribute any dataset as part of our code release.

Our implementation builds upon the GNN$^{+}$ codebase of \citet{luo2025can}, released under the MIT license and based in turn on GraphGPS \citep{rampavsek2022recipe}. The released code contains only our own implementation and requires users to download all datasets from their official sources.

\section{Sheaf-Specific Hyperparameters}

The message-passing block is the sheaf operator, parametrized along three axes: \textit{i)} the diffusion mechanism $\in \{\textsc{NSD}, \textsc{SAN}, \textsc{SANv2}\}$; \textit{ii)} the restriction-map family $\in \{\textsc{Diag}, \textsc{O}(d), \textsc{Gen}\}$, with $\textsc{O}(d)$ maps produced by Householder reflections; and \textit{iii)} the stalk dimension $d \in \{2,3,4\}$. The channels-per-stalk $f$ are chosen so that the per-node representation size $c = df$ does not exceed the reference width budget for each dataset. Following~\citet{bodnar2022neural}, restriction maps are predicted by a small learner with a bounded $\tanh$ activation, and the diffusion nonlinearity is $\sigma = \mathrm{elu}$. For the \textsc{Gen} parameterization, the block-diagonal inverse square root $D^{-1/2}$ is computed with a deterministic Newton--Schulz iteration rather than random-noise regularization to improve numerical stability.

\textsc{SAN} and \textsc{SANv2} use multi-head attention ($H = 8$) combined by channel-wise concatenation. The two models differ only in the attention scoring function: \textsc{SANv2} replaces the static GAT scoring of \textsc{SAN} with dynamic GATv2 scoring. The GNN$^{+}$ components toggled in the ablation are edge-feature conditioning, batch normalization, dropout, the residual connection, the post-message FFN (hidden multiplier $2$), and RWSE positional encoding.

\section{Complete Screening Results}

\begin{figure*}[htb!]
\centering
\includegraphics[width=\linewidth]{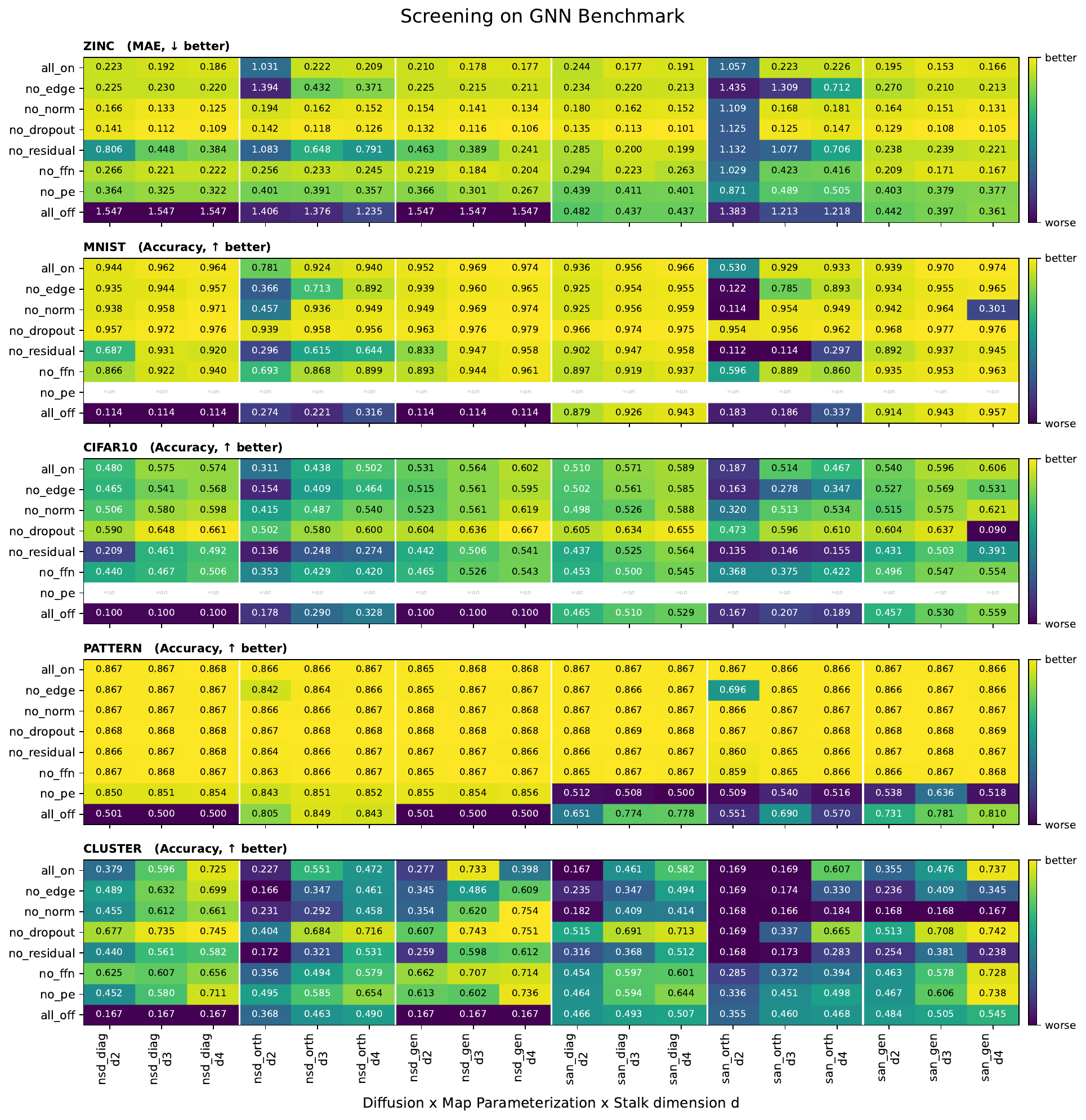}
\caption{Complete screening grid, GNN Benchmark. Rows are GNN$^{+}$ ablations, columns are the 18 architecture cells, color is within-dataset goodness (brighter is better, direction-aware), and the printed value is test-at-best-val.}
\label{fig:supp-heat-gnn}
\end{figure*}

\begin{figure*}[htb!]
\centering
\includegraphics[width=\linewidth]{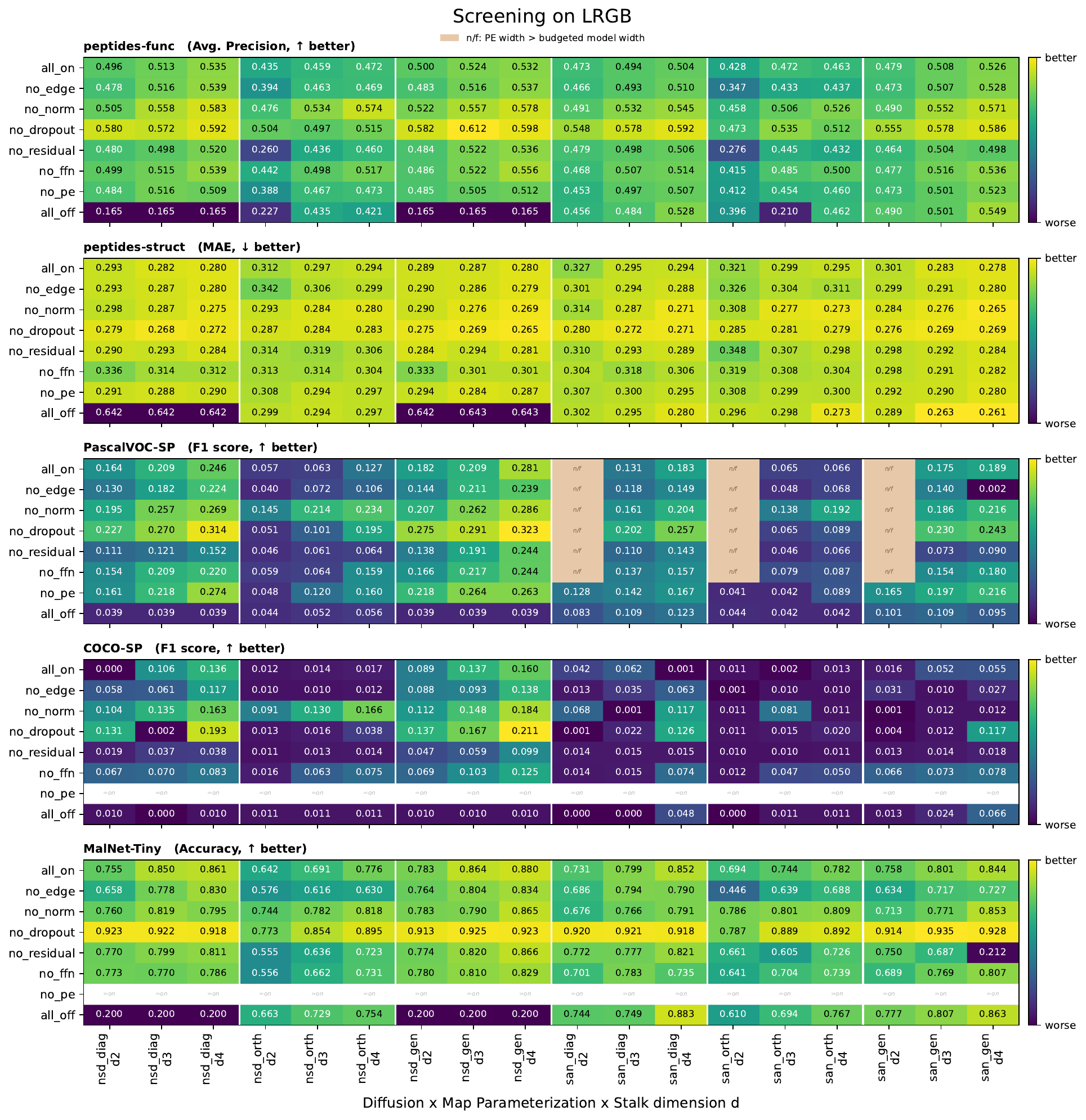}
\caption{Complete screening grid, LRGB.  Rows are GNN$^{+}$ ablations, columns are the 18 architecture cells, color is within-dataset goodness (brighter is better, direction-aware), and the printed value is test-at-best-val. The shaded \textsc{SAN} $d{=}2$ cells on PascalVOC-SP are infeasible at the parameter budget, as also reported in the main paper.}
\label{fig:supp-heat-lrgb}
\end{figure*}

\begin{figure*}[htb!]
\centering
\includegraphics[width=\linewidth]{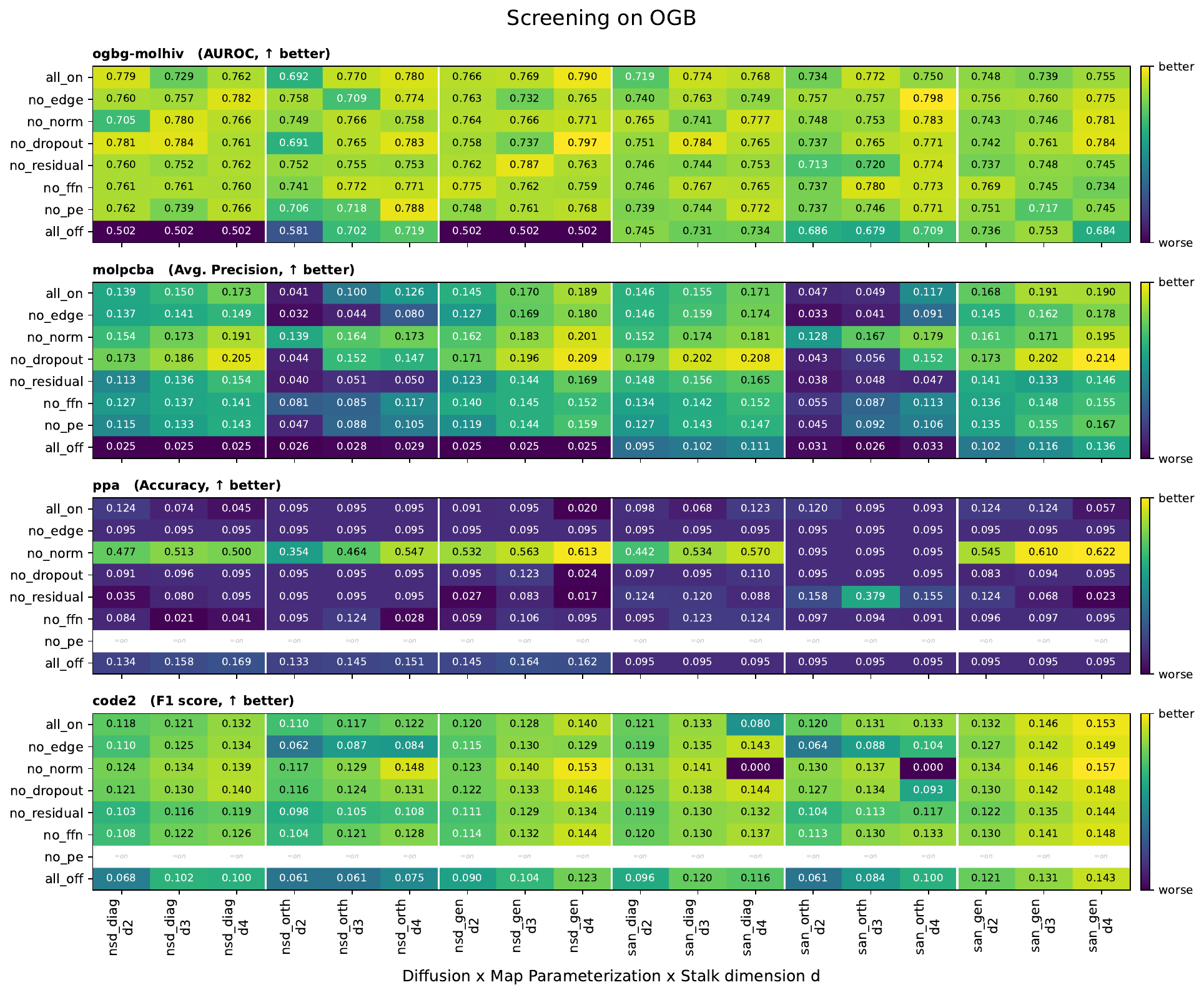}
\caption{Complete screening grid, OGB.  Rows are GNN$^{+}$ ablations, columns are the 18 architecture cells, color is within-dataset goodness (brighter is better, direction-aware), and the printed value is test-at-best-val. ogbg-ppa is the one dataset that systematically deviates from the trends seen elsewhere. The cause is the evaluation-time behavior of \texttt{BatchNorm}, discussed in the main paper.}
\label{fig:supp-heat-ogb}
\end{figure*}

Figures~\ref{fig:supp-heat-gnn}, \ref{fig:supp-heat-lrgb}, \ref{fig:supp-heat-ogb} report the complete screening grid for all 14 datasets: for each dataset, the 8 GNN$^{+}$ ablations are evaluated across the 18 architecture cells (diffusion mechanism $\times$ restriction-map parameterization $\times$ stalk dimension). Colors indicate within-dataset rank-normalized performance, with brighter colors corresponding to better-performing configurations regardless of the optimization direction, and each cell is annotated with the corresponding test-at-best-validation value. These figures report the 1,890 screening runs analyzed in the main paper.

On the six datasets that use no positional encoding, the \texttt{no\_pe} row is empty, since it coincides with the full recipe, and is marked \texttt{=on} rather than being run. The SAN cells at $d=2$ on PascalVOC-SP are infeasible at the dataset's parameter budget and are shaded separately.
The two positional-encoding-free ablations are unaffected and therefore run normally.

Reading the figures column-wise highlights the effect of the restriction-map parameterization: the $\textsc{O}(d)$ columns are visibly darker than the \textsc{Diag} and \textsc{Gen} columns on nearly every dataset, and the ordering is stable across both diffusion mechanisms and all three stalk dimensions.

Reading the figures row-wise highlights the contribution of the GNN$^{+}$ components: the \texttt{-dropout} row is brighter than the full recipe almost everywhere, whereas \texttt{-residual} and \texttt{all-removed} are consistently among the worst-performing configurations. \texttt{ogbg-ppa} is the only dataset that systematically deviates from these trends, for the reasons discussed in the main paper.

\section{Additional Confirmation Results}

The main paper reports 12 of the 14 datasets for conciseness. Table~\ref{tab:supp-confirm-full} completes the picture with COCO-SP and ogbg-ppa. Hyperparameter of confirmation sheaf models are reported in Table~\ref{tab:supp-confirm-configs}.

These results confirm what we observe in the main paper, with sheaf models unable to close the gap to the strongest baselines. On COCO-SP, a node-classification task, the gap is narrower, with NSD $0.10$ F1 points away from the best-performing method and leading the three mechanisms.
On the graph-level ogbg-ppa the gap widens to roughly $15$ accuracy points.
On both datasets, dynamic attention (SANv2) brings no visible benefit over static attention (SAN).

\end{document}